\documentclass[ams]{ISMM2007proceedings}

\newcommand{\fullpaperpdirectory}{.}
\newcommand{\fullpaperid}{xxx}

\usepackage[latin1]{inputenc}
\usepackage{algorithmic}
\usepackage{algorithm}

\RequirePackage[english]{babel}
\RequirePackage{tabularx}
\RequirePackage{graphicx}
\RequirePackage{subfigure}
\RequirePackage{hyperref}
\RequirePackage{amsrefs}

\begin{document}

%% A sample full paper template you modify and save as fullpaperdocument.tex
%% to produce your camera-ready full paper

%% +-------------+
%% | PAPER TITLE |
%% +-------------+

%\title[ short-title ]{ full-title }
\title%
    [Connections for hyperspectral image segmentation]% short title (maximum of 50 characters)
    {On distances, paths and connections for hyperspectral image segmentation}% full title

%% +---------+
%% | AUTHORS |
%% +---------+

\begin{Authors}
    \Author
        {Guillaume Noyel}
        \Affil
        [CMM]
        {Centre de Morphologie Mathématique, Ecole des Mines de Paris, 35 rue Saint-Honoré, 77305 Fontainebleau, France \\*
        \email{\{guillaume.noyel,jesus.angulo,dominique.jeulin\}@ensmp.fr}}% institutional e-mail address
    \Author
        {Jesus Angulo}
        \Affilref[CMM]
    \Author
        {Dominique Jeulin}
        \Affilref[CMM]
\end{Authors}

%% +-----------------------+
%% | ABSTRACT and KEYWORDS |
%% +-----------------------+
%% Abstract and keywords are written inside two environments,
%% the "Abstract" amd the "Keywords" environments, respectively.
%% Note the uppercase "A" and "K".
%% Keywords are separated with comma. The last keyword ends
%% with a period.
\begin{Abstract}
    The present paper introduces the $\eta$ and $\mu$ connections in order to add
    regional information on $\lambda$-flat zones, which only take into account
    a local information. A top-down approach is considered. First $\lambda$-flat zones are built in a way
    leading to a sub-segmentation. Then a finer segmentation is obtained by
    computing $\eta$-bounded regions and $\mu$-geodesic balls inside the $\lambda$-flat zones.
    The proposed algorithms for the construction of new
    partitions are based on queues with an ordered selection of seeds
    using the cumulative distance.
    $\eta$-bounded regions offers a control on the variations of amplitude in the class from a point, called center,
    and $\mu$-geodesic balls controls the ``size'' of the class. These results are
    applied to hyperspectral images.
\end{Abstract}


\section{Introduction}

The aim of this paper is to extend and to improve quasi-flat
zones-based segmentation. We focus on hyperspectral images to
illustrate our developments.

Flat zones, and its generalization, quasi-flat zones or
$\lambda$-flat zones, initially introduced for scalar functions
(i.e., gray-level images)~\cite{noyel:Meyer:1998} were generalized
to color (and multivariate) images~\cite{noyel:ZanogueraMeyer:2002}
(see definition \ref{\fullpaperid:Quasi-flat_zones} in section
\ref{\fullpaperid:sec:General notions}). Classically, $\lambda$-flat
zones are used as a method to obtain a first image partition (i.e.,
fine partition). The inconvenience is that the partition typically
presents small regions in zones of high gradient (e.g., close to the
contours, textured regions, etc.). Several previous works proposed
methods to solve it. In \cite{noyel:BrunnerSoille:2005}, Brunner and
Soille proposed a method to solve the over-segmentation of
quasi-flat zones. They use an iterative algorithm, on hyperspectral
images, based on seeds of area larger than a threshold. Their
approach consists in computing an over-segmentation and then merging
small regions. Besides, in \cite{noyel:SalembierGarridoGarcia:1998},
Salembier \emph{et al.} propose a method to suppress flat regions
with area below a given size. They work on the adjacency graph of
the flat zones and they define a merging order and criterion between
the regions. Moreover, a flat zone, below a given size, cannot be
shared by various flat zones. Crespo \emph{et al.}
\cite{noyel:CrespoSchafer:1997} proposed a similar approach for
region merging.

\begin{figure}[h]
    \centering
    \subfigure[]{\label{\fullpaperid:subfig:1a}\includegraphics[width=0.18\hsize]{\fullpaperpdirectory/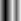}}
    \subfigure[]{\label{\fullpaperid:subfig:1b}\includegraphics[width=0.4\hsize]{\fullpaperpdirectory/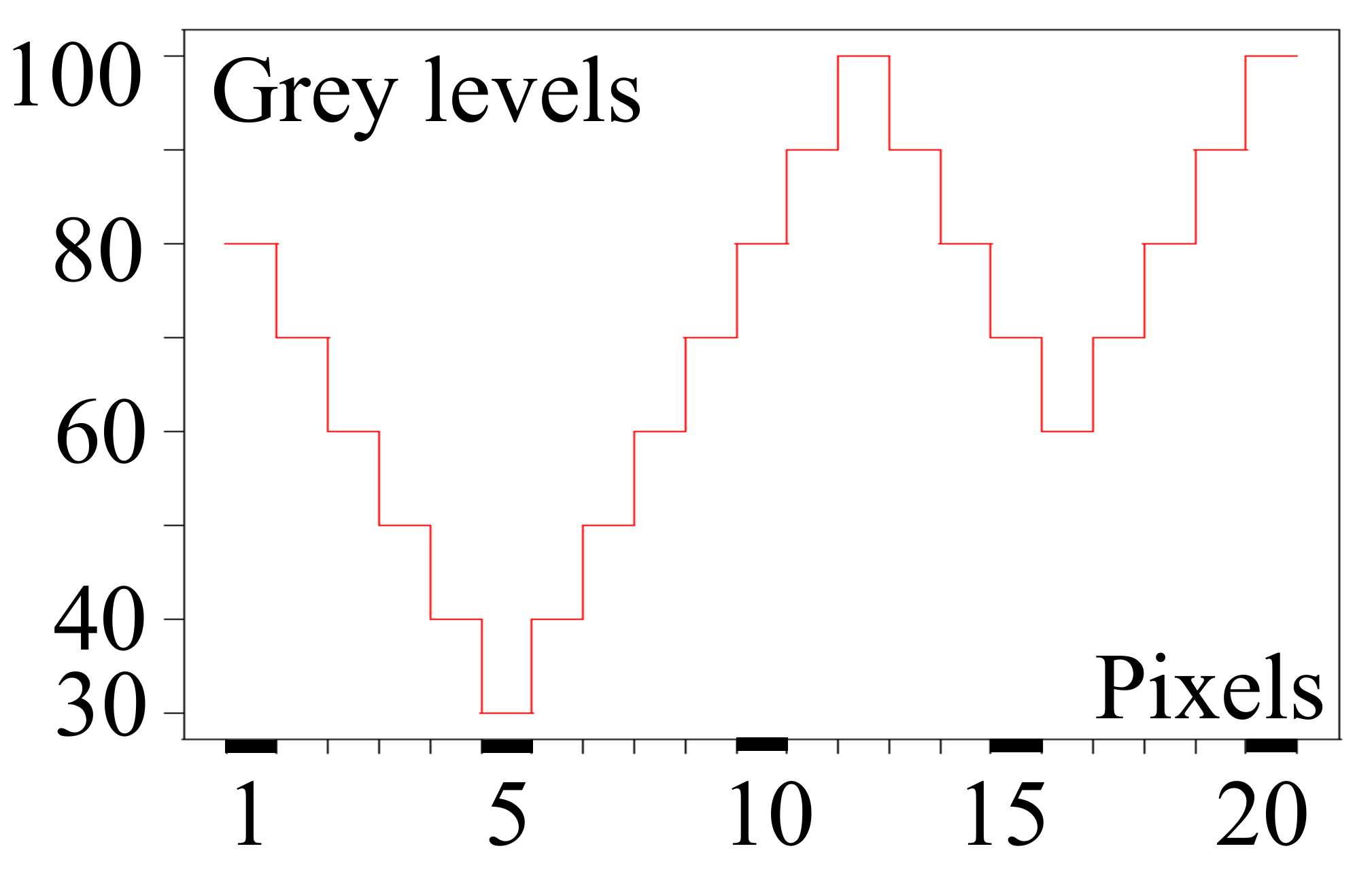}}
    \subfigure[]{\label{\fullpaperid:subfig:1c}\includegraphics[width=0.18\hsize]{\fullpaperpdirectory/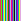}}
    \hspace{0.02\hsize}
    \subfigure[]{\label{\fullpaperid:subfig:1d}\includegraphics[width=0.18\hsize]{\fullpaperpdirectory/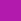}}
    \caption{$\lambda$-flat zones of image "tooth saw" ($21\times21$
    pixels). (a) Image. (b)~Row profile. (c)~$\lambda=9.9$ (21 zones). (d)~$\lambda=10$ (1 zone).}
    \label{\fullpaperid:lambda_FZ_im_tooth_saw}
\end{figure}

However, the main drawback of $\lambda$-flat zones for space
partition is that they are very sensitive to small variations of the
parameter $\lambda$. For instance in the example of
\autoref{\fullpaperid:lambda_FZ_im_tooth_saw}, we have only $21$ or
$1$ $\lambda$-flat zones depending on a slight variation of
$\lambda$. In fact, the problem of $\lambda$-flat zones lies in
their local definition: no regional information is taken into
account. For the example of
\autoref{\fullpaperid:lambda_FZ_im_tooth_saw}, with $\lambda=10$ is
obtained only one connected region, presenting a \emph{local smooth
variation} but involving a considerable \emph{regional rough
variation}. This problem is more difficult to tackle than the
suppression of small regions.

The purpose of our study is just to address this issue. We start
with an initial partition by $\lambda$-flat zones, with a non
critical high value of $\lambda$, that leads to a sub-segmentation
(i.e., large classes in the partition). Then, for each class, we
would like to define a second segmentation according to a regional
criterion. In fact, two new connections are introduced: (1)
$\eta$-bounded regions, and (2) $\mu$-geodesic balls; the
corresponding algorithms are founded on seed-based region growing
inside the $\lambda$-flat zones. We show that the obtained reliable
segmentations are less critical with respect to the choice of
parameters and that these new segmentation approaches are
appropriate for hyperspectral images. From a more theoretical
viewpoint, the Serra's theory of
segmentation~\cite{noyel:Serra:2005} allows us to explain many
notions which are considered in this paper.

\section{General notions}
\label{\fullpaperid:sec:General notions}

In this section some notions necessary for the sequel are reminded.

Hyperspectral images are multivariate discrete functions with
typically several tens or hundreds of spectral bands. In a formal
way, each pixel of an hyperspectral image is a vector with values in
wavelength, in time, or associated with any index $j$. To each
wavelength, time or index corresponds an image in two dimensions,
called channel.

The number of channels depends on the nature of the specific problem
under study (satellite imaging, spectroscopic images, temporal
series, etc.).

\begin{definition}[Hyperspectral image]
    \label{\fullpaperid:Hyperspectral_image}% only one word
Let $\mathbf{f_{\lambda}}:E \rightarrow \mathcal{T}^{L}$ ($x
\rightarrow \mathbf{f}_{\mathbf{\lambda}}(x) = $ $\left(
f_{\lambda_{1}}(x), f_{\lambda_{2}}(x), \ldots, f_{\lambda_{L}}(x)
\right)$), be an hyperspectral image, where: $E \subset
\mathbb{R}^{2}$, $\mathcal{T} \subset \mathbb{R}$ and
$\mathcal{T}^{L} = $ $\mathcal{T} \times \mathcal{T} \times \ldots
\times \mathcal{T}$; $x = x_{i} \ \backslash \ i\in\{1,2, \ldots, P
\}$ is the spatial coordinates of a vector pixel
$\mathbf{f}_{\lambda}(x_{i})$ ($P$ is the pixels number of $E$);
$f_{\lambda_{j}} \ \backslash \ j \in \{1,2, \ldots, L\}$ is a
channel ($L$ is the channels number); $f_{\lambda_{j}}(x_{i})$ is
the value of vector pixel $\mathbf{f}_{\lambda}(x_{i})$ on channel
$f_{\lambda_{j}}$.
\end{definition}

\begin{definition}[Spectral distance]% between square brackets
    \label{\fullpaperid:spectral_distance}% only one word
A spectral distance is a function $d: \mathcal{T}^{L} \times
\mathcal{T}^{L} \rightarrow \mathbb{R}^{+}$ which verifies the
properties: 1) $d(\mathbf{s}_{i},\mathbf{s}_{j})\geq 0$,
2)$d(\mathbf{s}_{i},\mathbf{s}_{j}) = 0$ $\Leftrightarrow
\mathbf{s}_{i}=\mathbf{s}_{j}$,
3)$d(\mathbf{s}_{i},\mathbf{s}_{j})=$
$d(\mathbf{s}_{j},\mathbf{s}_{i})$,
4)$d(\mathbf{s}_{i},\mathbf{s}_{j})\leq $
$d(\mathbf{s}_{i},\mathbf{s}_{k})+$
$d(\mathbf{s}_{k},\mathbf{s}_{j})$, for $\mathbf{s}_{i}$,
$\mathbf{s}_{j}$, $\mathbf{s}_{k} \in \mathcal{T}^{L}$
\end{definition}

Various metrics distance are useful for hyperspectral points. In
this paper, the following two are used:
\begin{itemize}
  \item Euclidean distance:
    $ d_{E}(\mathbf{f_{\lambda}}(x), \mathbf{f_{\lambda}}(y)) =
    \sqrt{ \sum_{j=1}^{L}( f_{\lambda_{j}}(x) - f_{\lambda_{j}}(y) )^2
    }$
  \item Chi-squared distance:\\
    $d_{\chi^{2}}( \mathbf{f}_{\lambda}(x_{i}) ,
    \mathbf{f}_{\lambda}(x_{i'}) ) = \sqrt{\sum_{j=1}^{L}
    \frac{N}{f_{.\lambda_{j}}} \left( \frac{ f_{\lambda_{j}}(x_{i}) }{
    f_{x_{i}.} } - \frac{ f_{\lambda_{j}}(x_{i'}) }{ f_{x_{i'}.} }
    \right)^{2}}$
with $f_{.\lambda_{j}} = \sum_{i=1}^{P} f_{\lambda_{j}}(x_{i})$,
$f_{x_{i}.} = \sum_{j=1}^{J} f_{\lambda_{j}}(x_{i})$ and $N =
\sum_{j=1}^{L}\sum_{i=1}^{P}f_{\lambda_{j}}(x_{i})$.
\end{itemize}

\begin{definition}[Path]
    \label{\fullpaperid:path}
    A path between two points $x$ and $y$  is a chain of points
    $\left(p_0, p_1, \ldots, p_i, \ldots, p_n \right) \in E$ such as
    $p_0 = x$ and $p_n=y$, and for all $i$, $(p_i, p_{i+1})$ are
    neighbours.
\end{definition}

\begin{definition}[Quasi-flat zones or $\lambda$-flat zones]% between square brackets
    \label{\fullpaperid:Quasi-flat_zones}% only one word
Given a distance $d : \mathcal{T}^{L} \times \mathcal{T}^{L}
\rightarrow \mathbb{R}^{+}$, two points $x$, $y$ $\in E$ belongs to
the same quasi-flat zone of an hyperspectral image
$\mathbf{f}_{\lambda}$ if and only if there is a path $(p_0,p_1,
\ldots, p_n) \in E^n$ such as $p_0=x$ and $p_n=y$ and, if, for all
$i$, $(p_i,p_{i+1})\in E^2$ are neighbours and
$d\left(\mathbf{f}_{\lambda}(p_i), \mathbf{f}_{\lambda}(p_{i+1})
\right) \leq \lambda$, with $\lambda \in \mathbb{R}^{+}$.
\end{definition}

A path $(p_0,p_1, \ldots, p_n)$ in an hyperspectral image can be
seen as a graph in which the nodes correspond to the points
connected by edges along the path. For all $i$, the edge between the
nodes $p_i$ and $p_{i+1}$ is weighted by
$d\left(\mathbf{f}_{\lambda}(p_i), \mathbf{f}_{\lambda}(p_{i+1})
\right)$.

\begin{definition}[Geodesic path]% between square brackets
    \label{\fullpaperid:Geodesic_path}% only one word
The geodesic path between two points $x$ and $y$ in $E$ is the path
of minimum weight.
\end{definition}

This definition means that the sum of distances
$\sum_{i=0}^{n}d\left(\mathbf{f}_{\lambda}(p_i),
\mathbf{f}_{\lambda}(p_{i+1}) \right)$, along this path, $(p_0,p_1,
\ldots, p_n)$ such as $p_0=x$ and $p_n=y$, is minimum. It is called
the geodesic distance between $x$ and $y$ and noted $d_{geo}(x,y)$.

In order to compute the geodesic path, the Dijkstra's algorithm can
be used~\cite{noyel:Dijkstra:1959}. Meanwhile we use an algorithm
based on hierarchical queues ~\cite{noyel:Vincent:1998}.

For the purposes of segmentation, we need to fix some theoretical
notions.

\begin{definition}[Partition]
    \label{\fullpaperid:Partition}
Let $E$ be an arbitrary set. A partition ${\cal D}$ of $E$ is a
mapping $x \rightarrow D(x)$ from $E$ into ${\cal P}(E)$ such that:
(i) for all $x\in E$: $x \in D(x)$, (ii) for all $x,y \in E$:
$D(x)=D(y)$ or $D(x) \cap D(y) = \emptyset$. $D(x)$ is called the
class of the partition of origin $x$.
\end{definition}

The set of partitions of an arbitrary set $E$ is ordered as follows.

\begin{definition}[Order of partitions]
    \label{\fullpaperid:Order_partitions}
A partition ${\cal A}$ is said to be finer (resp. coarser) than a
partition ${\cal B}$, $\cal A \leq \cal B$ (resp. $\cal A \geq \cal
B$), when each class of ${\cal A}$ is included in a class of ${\cal
B}$.
\end{definition}

This leads to the notion of ordered hierarchy of partitions
$\Pi_{i=1}^{N} {\cal D}_{i}$, such that ${\cal D}_{i} \leq {\cal
D}_{i+1}$, and even to a complete lattice~\cite{noyel:Serra:2005}.

\begin{definition}[Connection]
    \label{\fullpaperid:Connection}
Let $E$ be an arbitrary non empty set. We call connected class or
connection ${\cal C}$ any family in ${\cal P}(E)$ such that: (0)
$\emptyset \in {\cal C}$, (i) for all $x\in E$, $\{x\} \in {\cal
C}$, (ii) for each family ${C_{i}, i\in I}$ in ${\cal C}$,
$\cap_{i}C_{i} \neq \emptyset$ implies $\cup_{i}C_{i} \in {\cal C}$.
Any set $C$ of a connected class ${\cal C}$ is said to be connected.
\end{definition}

It is clear that a connection involves a partition, and consequently
a segmentation of $E$. According to~\cite{noyel:Serra:2005}, more
precise notions than connective criteria (which produce
segmentations) can be considered in order to formalize the theory,
but this is out of the scope of this paper.

In particular, the $\lambda$ flat zones can be considered as a
connection, $\lambda$-flat connection, i.e., $\lambda FZ$ is the
partition of the image $\mathbf{f}_{\lambda}$ according to the
$\lambda$-flat connection.

For multivariate images, where the extrema (i.e., minimum or
maximum) are not defined, the vectorial median is a very interesting
notion to rank and select the points
\cite{noyel:AstolaHaavisto:1990}.

\begin{definition}[Vectorial median]
    \label{\fullpaperid:vectorial_median}
A vectorial median of a set $R \subset E$ is any value
$\mathbf{f}_{\lambda}(k)$ in the set at point $k \in R$ such as:
\begin{equation}\label{\fullpaperid:eq_vectorial_median}
    k = argmin_{p\in R} \sum_{i / x_i \in R} d\left(\mathbf{f}_{\lambda}(p),
    \mathbf{f}_{\lambda}(x_{i})\right) = argmin_{p\in R} \delta_R(\mathbf{f}_{\lambda}(p))
\end{equation}
\end{definition}

In order to compute the vectorial median $\mathbf{f}_{\lambda}(k)$,
$k \in R$, the cumulative distance
$\delta_R\left(\mathbf{f}_{\lambda}(p)\right)$ has to be evaluated
for all $p \in R$. Therefore all points $p$ are sorted, in ascending
order, on the cumulative distance. The resulting list of ordered
points is called ascending ordered list based on the cumulative
distance. The first element of the list is the vectorial median (the
last element is considered as the anti-median).

\section{$\eta$-bounded regions}

One of the main idea is to understand that on quasi-flat zones the
distance or slope between two neighbouring points must be inferior
to the parameter $\lambda$. We can establish a comparison with a
hiker, from one point who will only deal with the local slope and
not on the cumulative difference in altitude on the $\lambda$-flat
zones. To consider this limitation we propose a new kind of regional
zones: $\eta$-bounded regions, according to the following
connection.

\begin{definition}[$\eta$-bounded connection]
    \label{\fullpaperid:Eta_Connection}
Given an hyperspectral image $\mathbf{f}_{\lambda}(x)$ and its
initial partition based on $\lambda$-flat zones, $\lambda FZ$, where
$\lambda FZ_{i}$ is the connected class $i$ and $R_{i}\subseteq E$
(with cardinal $K$) is the set of points $p_{k}$, $k={0,1,2, ... ,
K-1}$, that belongs to the class $i$. Let $p_{0}$ be a point of
$R_{i}$, named the center of class $i$, and let $\eta\in
\mathbb{R}^+$ be a positive value. A point $p_k$ belongs to the
$\eta$-connected component centered at $p_{0}$, denoted $\eta
BR_{i}^{p_{0}}$, if and only if
$d(\mathbf{f}_{\lambda}(p_{0}),\mathbf{f}_{\lambda}(p_{k}))\leq
\eta$ and $p_{0}$ and $p_{k}$ are connected.
\end{definition}

For each class $\lambda FZ_{i}$ the method is iterated using
different centers $p_{j}$ ($j=0,1,\cdots J$) until all the space of
the $\lambda$-connected component is segmented: $\cup_{j=0}^{J} \eta
BR_{i}^{p_{j}}=\lambda FZ_{i}$, $\cap_{j=0}^{J} \eta
BR_{i}^{p_{j}}=\emptyset$, where the $\eta$-bounded regions are also
connected. Each center $p_{j}$ belongs to $\lambda FZ_{i} \setminus
\cup_{l=0}^{j-1} \eta BR_{i}^{p_{l}}$.

The new image partition associated to $\eta$-bounded connection is
denoted $\eta BR$. It is evident that this second-class connection
is contained in the $\lambda FZ$ initial connection, i.e., $\eta
BR(x) \leq \lambda FZ(x)$. As shown by
Serra~\cite{noyel:Serra:2005}, a center or seed $p_{j}$ is needed to
guarantee the connectiveness, thus being precise, the region is
bounded with respect to the center. The $\eta$-connection is a
generalization of the jump connection~\cite{noyel:Serra:1999} with
the difference that working on hyperspectral images, the seeds
$p_{j}$ cannot be the minima or maxima. We propose to compute the
median value as initial seed.

Note that the method can be also applied on the space $E$ of the
initial image $\mathbf{f}_{\lambda}$, without considering an initial
$\lambda FZ$ (which is equivalent to take a value of $\lambda$ equal
to the maximal image distance range). The advantage of our approach
is that we have now a control of the local variation, limited by
$\lambda$, and the regional variation, bounded by $\eta$. Moreover,
the computation of seeds is more coherent when working on relatively
homogenous regions.

In practice, for all the $K$ points on each $\lambda FZ_{i}$, the
ascending ordered list based on cumulative distance $\delta_i$ is
computed. Then, the first element of the list, i.e. the vectorial
median, is used as first seed $k$ of the $\eta$-bounded region $\eta
BR_{i}^{1}$. The distance, from the seed to each point $p \in
\lambda FZ_{i}$, such as $p \in Neigh(q)$, $q \in \eta BR_{i}^{1}$,
is measured. If this distance is less than $\eta$ then $p$ is added
to $\eta BR_{i}^{1}$ and removed from $\delta_i$. For all the others
points $q \in \lambda FZ_{i} \setminus \eta BR_{i}^{1}$, the first
point of the list $\delta_i$ is the seed of the second
$\eta$-bounded region $\eta BR_{i}^{2}$. Then we iterate the process
until all the points of the $\lambda$-flat zone  $\lambda FZ_{i}$
are in an $\eta$-bounded region. %(algorithm
%\ref{\fullpaperid:algetaBR}).

\begin{algorithm}[ht]
\small{\caption{$\mu$-geodesic balls}
\label{\fullpaperid:algetaBR}
\algsetup{indent =2em}
\begin{algorithmic}
\STATE Given a distance $d$, the $\lambda$-flat zones, $\lambda FZ$, of an image $\mathbf{f}_{\lambda}$, $\delta$ a list of cumulative distance, $Q$ a queue, $imOut$ an output scalar image \\
\STATE Initialize the value of $\eta$\\
\STATE $currentlabel \leftarrow 0$\\
\FORALL{$\lambda FZ \in \mathbf{f}_{\lambda}$ }
    \FORALL{point $p \in \lambda FZ$}
        \STATE $distance \leftarrow \sum_{q \in \lambda FZ}d\left(p,q \right)$
        \STATE $\delta \leftarrow$ add the pair $(p, distance)$
    \ENDFOR
    \STATE Ascending sort on parameter $distance$ of $\delta$
    \WHILE{$\delta$ is not empty}
        \STATE $k \leftarrow$ first point of $\delta$
        \STATE $Q \leftarrow push(k)$
        \WHILE{$Q$ is not empty}
            \STATE $p \leftarrow pop(Q)$
            \STATE $imOut(p) \leftarrow currentlabel$
            \STATE Remove $q$ and its $distance$ in $\delta$
            \FORALL{$q \in Neigh(p)$ and $q \in \delta$}
                \IF{$d\left(k,q \right) \leq \eta$}
                    \STATE $Q \leftarrow push(q)$
                \ENDIF
            \ENDFOR
        \ENDWHILE
        \STATE $currentlabel \leftarrow currentlabel + 1$
    \ENDWHILE
\ENDFOR
\end{algorithmic}}
\end{algorithm}

\pagebreak
Using again the mountain comparison, this can be compared
to a hiker starting from a point with a walk restricted to a ball of
diameter $2 \times \eta$ centered on the starting point. He cannot
go upper or lower than this boundary. The points to be reached by
the hiker, given the previous conditions, are connected and
constitute an $\eta$-bounded region.

\begin{figure}[!ht]
    \centering
    \subfigure[]{\label{\fullpaperid:subfig:1a}\includegraphics[width=0.18\hsize]{\fullpaperpdirectory/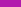}}
    \hspace{0.005\hsize}
    \subfigure[]{\label{\fullpaperid:subfig:1b}\includegraphics[width=0.18\hsize]{\fullpaperpdirectory/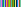}}
    \hspace{0.005\hsize}
    \subfigure[]{\label{\fullpaperid:subfig:1c}\includegraphics[width=0.18\hsize]{\fullpaperpdirectory/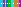}}
    \hspace{0.005\hsize}
    \subfigure[]{\label{\fullpaperid:subfig:1d}\includegraphics[width=0.18\hsize]{\fullpaperpdirectory/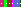}}
    \hspace{0.005\hsize}
    \subfigure[]{\label{\fullpaperid:subfig:1e}\includegraphics[width=0.18\hsize]{\fullpaperpdirectory/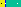}}

    \subfigure[]{\label{\fullpaperid:subfig:1f}\includegraphics[width=0.18\hsize]{\fullpaperpdirectory/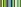}}
    \hspace{0.005\hsize}
    \subfigure[]{\label{\fullpaperid:subfig:1g}\includegraphics[width=0.18\hsize]{\fullpaperpdirectory/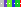}}
    \hspace{0.005\hsize}
    \subfigure[]{\label{\fullpaperid:subfig:1h}\includegraphics[width=0.18\hsize]{\fullpaperpdirectory/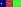}}
    \hspace{0.005\hsize}
    \subfigure[]{\label{\fullpaperid:subfig:1i}\includegraphics[width=0.18\hsize]{\fullpaperpdirectory/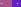}}
    \caption{$\eta$-bounded regions and $\mu$-geodesic balls of image "tooth saw" ($21\times21\times4$ pixels)
    for $\lambda = 10$. Seeds are marked with a white (or black) point, when they
    are not trivial. (a)~$\lambda FZ$. $\lambda=10$. (b)~$\eta = 0$. (c)~$\eta = 10$. (d)~$\eta=20$.
    (e)~$\eta=30$. (f)~$\mu = 0$. (g)~$\mu = 20$. (h)~$\mu=40$. (i)~$\mu=100$.}
    \label{\fullpaperid:fig_eta_BR_mu_GB_im_tooth_saw}
\end{figure}

To understand the effect of these regions, they are applied to the
image "tooth saw" with an Euclidean distance $d_E$
(\autoref{\fullpaperid:fig_eta_BR_mu_GB_im_tooth_saw}). For the sake
of simplicity, in this image only the grey levels of the first
channel have a shape of "tooth saw"
(\autoref{\fullpaperid:lambda_FZ_im_tooth_saw} (b)), the others
being constant.

\begin{figure}[!ht]
    \centering
    \subfigure[]{\label{\fullpaperid:subfig:1a}\includegraphics[width=0.4\hsize]{\fullpaperpdirectory/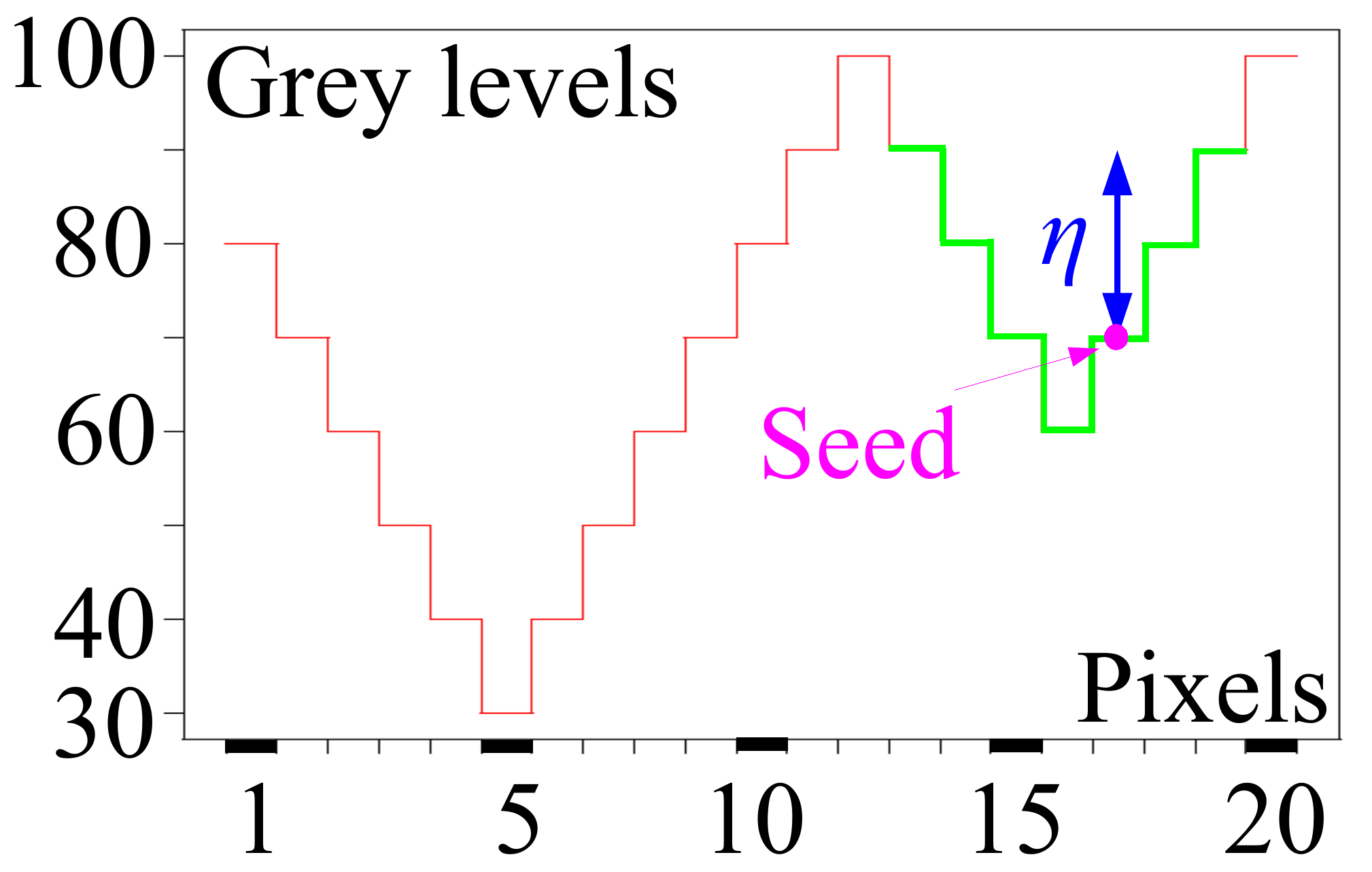}}
    \hspace{0.1\hsize}
    \subfigure[]{\label{\fullpaperid:subfig:1b}\includegraphics[width=0.4\hsize]{\fullpaperpdirectory/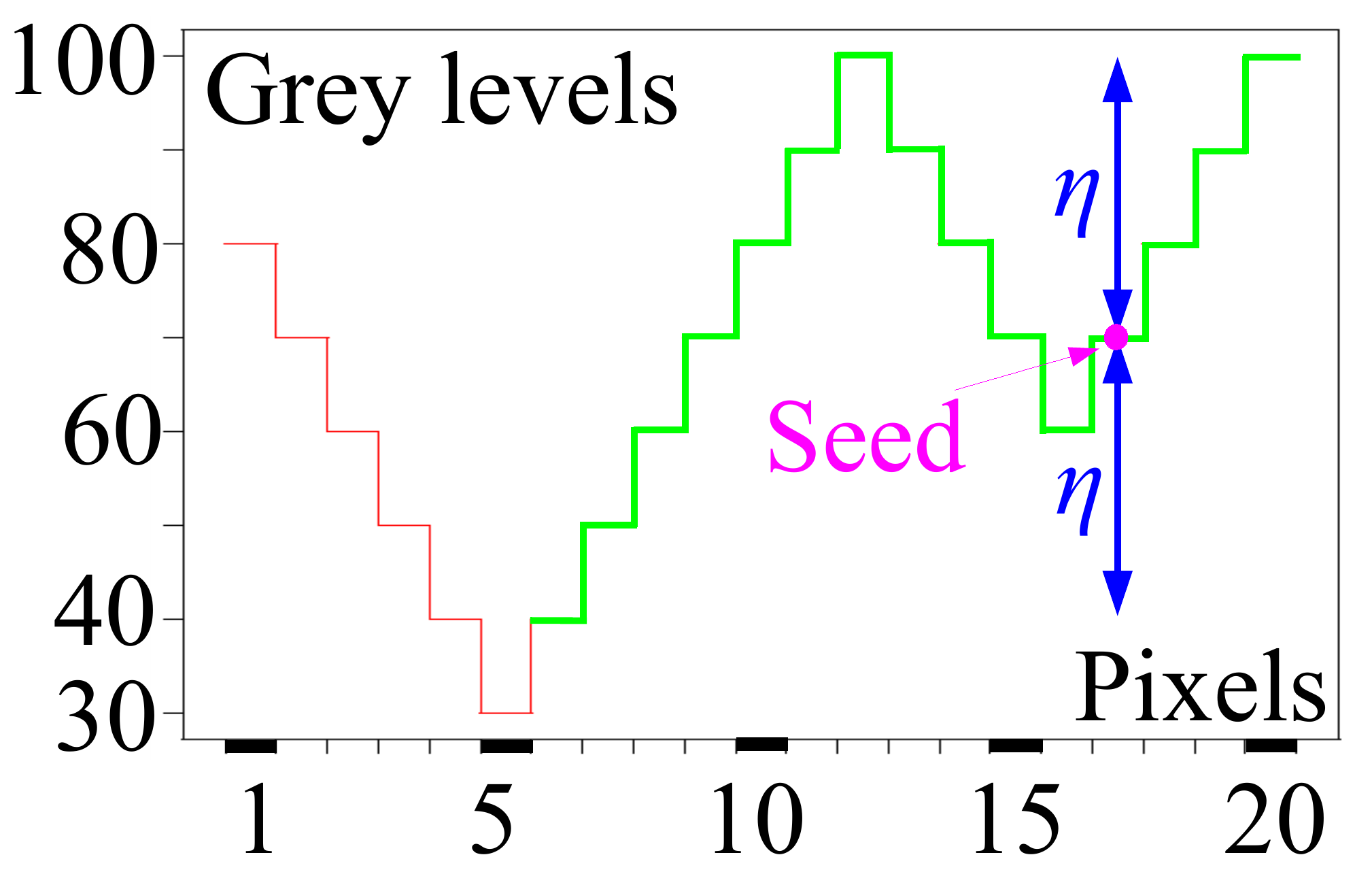}}
    \subfigure[]{\label{\fullpaperid:subfig:1c}\includegraphics[width=0.4\hsize]{\fullpaperpdirectory/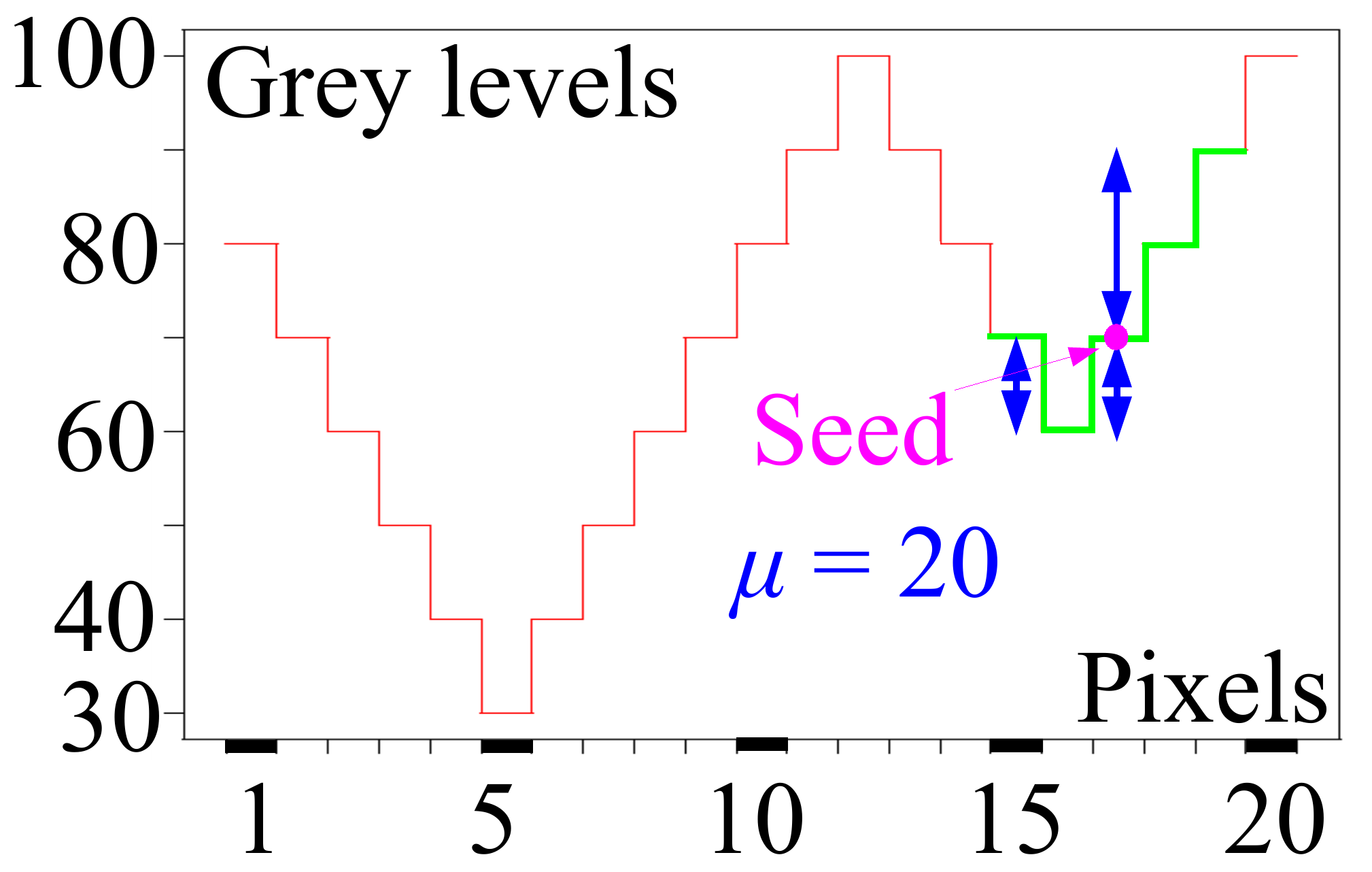}}
    \hspace{0.1\hsize}
    \subfigure[]{\label{\fullpaperid:subfig:1d}\includegraphics[width=0.4\hsize]{\fullpaperpdirectory/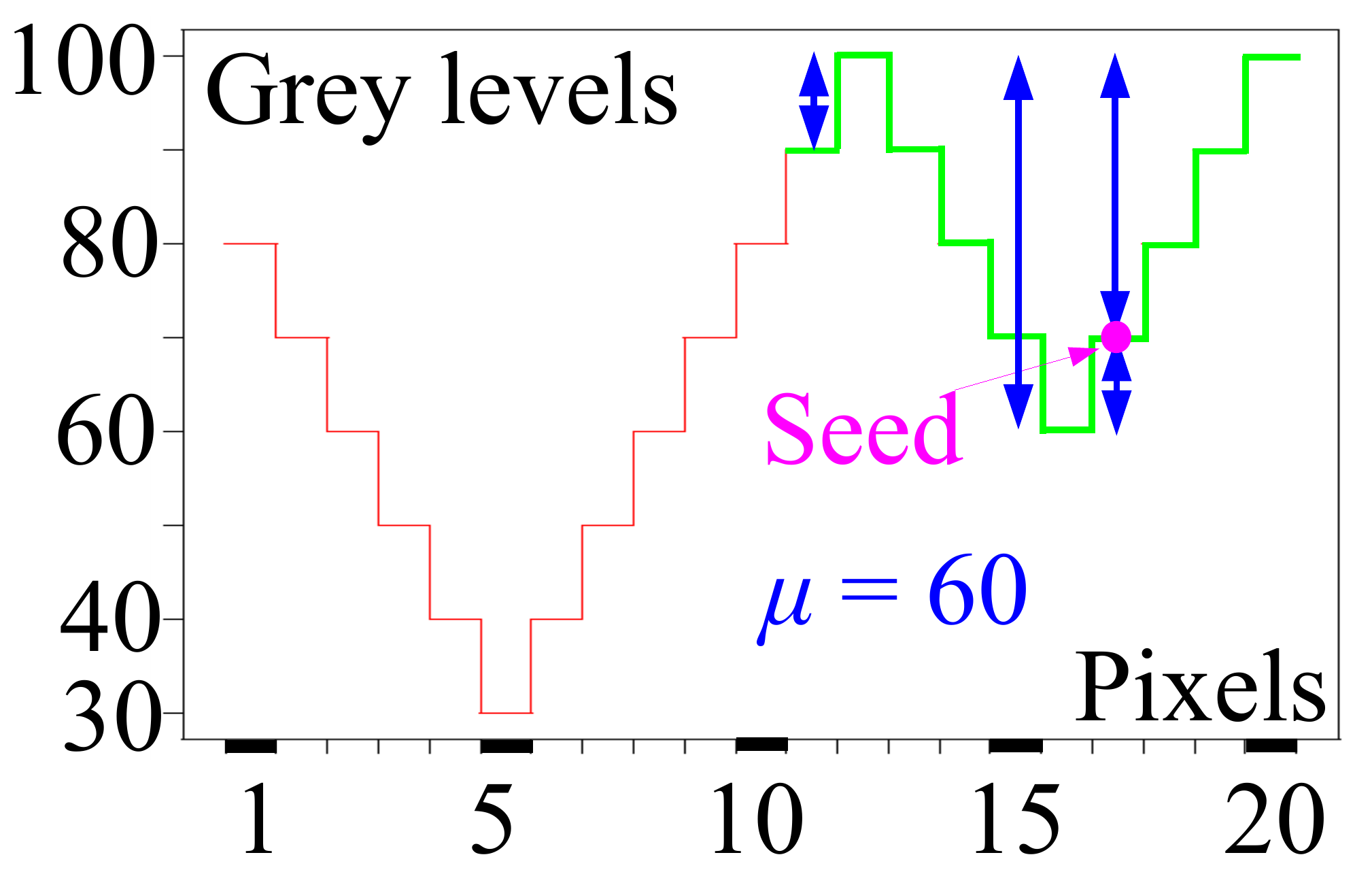}}
    \caption{$\eta$-bounded region and $\mu$-geodesic ball (in green) on the profile "tooth saw" for $\lambda =
    10$. (a)~$\eta = 20. $ (b)~$\eta = 30$. (c)~$\mu = 20$. (d)~$\mu = 60$.}
    \label{\fullpaperid:fig_eta_BR_mu_GB_profils}
\end{figure}

We notice from the figure that the smallest the parameter $\eta$,
the smallest the area of $\eta$-bounded regions. Besides,
$\eta$-bounded regions are very sensitive to the peaks when working
on scalar functions. In fact, if $\eta$ is less than the difference
of altitude between the seed and the peak, the points in between are
in the same $\eta$-bounded region. However, if $\eta$ is larger than
the difference of altitude between the seed and the peak, the points
behind the peak, for which the difference of altitude from the seed
is less than $\eta$, are in the same $\eta$-bounded region
(\autoref{\fullpaperid:fig_eta_BR_mu_GB_profils}).

%\clearpage
\section{$\mu$-geodesic balls}

As for $\eta$-bounded regions, we have created $\mu$-geodesic balls,
$\mu GB$, to improve the $\lambda$-flat zones, but now introducing a
control of the dimension of the zone. First of all, $\lambda$-flat
zones are built. Then, the cumulative difference in altitude is
measured from a starting point, the seed, in each $\lambda$-flat
zone. From this point a geodesic ball of radius $\mu$ is computed.
This zone is a $\mu$-geodesic ball. It corresponds to the maximum of
steps (points) that can be reached by a hiker starting from the seed
inside a $\lambda$-flat zone, for a given cumulative altitude.
$\mu$-geodesic balls are defined using the following connection:

\begin{definition}[$\mu$-geodesic connection]
    \label{\fullpaperid:Mu_Connection}
Given an hyperspectral image $\mathbf{f}_{\lambda}(x)$ and its
initial partition based on $\lambda$-flat zones, $\lambda FZ$, where
$\lambda FZ_{i}$ is the connected class $i$ and $R_{i}\subseteq E$
(with cardinal $K$) is the set of points $p_{k}$, $k={0,1,2, ... ,
K-1}$, that belongs to the class $i$. Let $p_{0}$ be a point of
$R_{i}$, named the center of class $i$, and let $\eta\in
\mathbb{R}^+$ be a positive value. A point $p_k$ belongs to the
$\mu$-connected component centered at $p_{0}$, denoted $\mu
GB_{i}^{p_{0}}$ if and only if
$d_{geo}(\mathbf{f}_{\lambda}(p_{0}),\mathbf{f}_{\lambda}(p_{k}))\leq
\mu$.
\end{definition}

It is important to notice, that the geodesic paths imposed the
connectivity to the $\mu$-geodesic ball. Formally, for each class
$\lambda FZ_{i}$ the method is iterated using different centers
$p_{j}$ ($j=0,1,\cdots J$) until the full segmentation of the
$\lambda$-connected component.

\begin{algorithm}[h]
\small{\caption{$\mu$-geodesic balls} \label{\fullpaperid:algmuRZ}
\algsetup{indent =2em}
\begin{algorithmic}
 \STATE Given a distance $d$, the $\lambda$-flat zones, $\lambda FZ$, of an image
$\mathbf{f}_{\lambda}$, $\delta$ a list of cumulative distance, $Q$ a queue, $imOut$ an output scalar image \\
\STATE Initialize the value of $\mu$\\
\STATE $currentlabel \leftarrow 0$\\
\FORALL{$\lambda FZ \in \mathbf{f}_{\lambda}$ }
    \FORALL{point $p \in \lambda FZ$}
        \STATE $distance \leftarrow \sum_{q \in \lambda FZ}d\left(p,q
        \right)$
        \STATE $\delta \leftarrow$ add the pair $(p, distance)$
    \ENDFOR
    \STATE Ascending sort on parameter $distance$ of $\delta$
    \WHILE{$\delta$ is not empty}
        \STATE $k \leftarrow$ first point of $\delta$
        \FORALL{point $p$ inside the geodesic ball of center $k$ and radius $\mu$ inside the $\lambda FZ$}
            \STATE $imOut(p) \leftarrow currentlabel$
            \STATE Remove $p$ and its $distance$ in $\delta$
        \ENDFOR
        \STATE $currentlabel \leftarrow currentlabel + 1$
    \ENDWHILE
\ENDFOR
\end{algorithmic}}
\end{algorithm}

The new image partition associated to $\mu$-geodesic connection is
denoted $\mu GB$. As for $\eta$-bounded connection, this
second-class connection is contained in the $\lambda FZ$ initial
connection, i.e., $\mu GB(x) \leq \lambda FZ(x)$. The advantage of
this approach is that we have now a regional control of the
``geodesic size'' of the classes by measuring the geodesic distance,
limited by $\mu$ inside the local variation limited by $\lambda$. In
practice, $\mu$-geodesic balls are built as $\eta$-bounded regions,
except that from each seed the geodesic ball is computed inside the
$\lambda FZ$.

$\mu$-geodesic balls are computed with an Euclidean distance $d_E$
for the image "tooth saw"
(\autoref{\fullpaperid:fig_eta_BR_mu_GB_im_tooth_saw}). We notice
that the smaller is $\mu$, the smaller the area of $\mu$-geodesic
balls is. Besides, these balls are not very sensitive to the peaks.
In fact, if $\mu$ is less than the difference of altitude between
the seed and the peak, the points in between are in the same
$\mu$-geodesic ball. However, if $\mu$ is larger than the difference
of altitude between the seed and the peak, only the points behind
the peak, for which the cumulative altitude from the seed is less
than $\mu$, are in the same $\mu$-geodesic ball
(\autoref{\fullpaperid:fig_eta_BR_mu_GB_profils}).

\section{Results and discussions}

\begin{figure}[h]
    \centering
    \subfigure[]{\label{\fullpaperid:subfig:1a}\includegraphics[width=0.18\hsize]{\fullpaperpdirectory/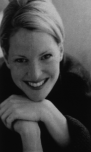}}
    \hspace{0.005\hsize}
    \subfigure[]{\label{\fullpaperid:subfig:1b}\includegraphics[width=0.18\hsize]{\fullpaperpdirectory/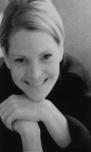}}
    \hspace{0.005\hsize}
    \subfigure[]{\label{\fullpaperid:subfig:1c}\includegraphics[width=0.18\hsize]{\fullpaperpdirectory/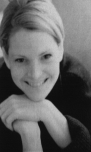}}
    \hspace{0.005\hsize}
    \subfigure[]{\label{\fullpaperid:subfig:1d}\includegraphics[width=0.18\hsize]{\fullpaperpdirectory/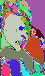}}
    \hspace{0.005\hsize}
    \subfigure[]{\label{\fullpaperid:subfig:1e}\includegraphics[width=0.18\hsize]{\fullpaperpdirectory/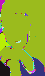}}
    \caption{Channels of image "woman face" $\mathbf{f}_{\lambda}$ ($45 \times 76 \times 61$
    pixels) and $\lambda$-flat zones.
    (a)~ $\mathbf{f}_{\lambda_{30}}$. (b)~$\mathbf{f}_{\lambda_{45}}$.
    (c)~ $\mathbf{f}_{\lambda_{61}}$. (d)~ $\lambda=0.003$.
    (e)~ $\lambda=0.006$.
    (Source: Spectral Database, University of Joensuu Color Group,
    \url{http://spectral.joensuu.fi/})}
    \label{\fullpaperid:fig_im_woman_face}
\end{figure}

In order to illustrate our results on real images, we extracted
$\eta$-bounded regions and $\mu$-geodesic balls in the image "woman
face" of size $45 \times 76 \times 61$ pixels
(\autoref{\fullpaperid:fig_im_woman_face}). The channels are
acquired between 400 nm and 700 nm with a step of 5 nm. Moreover, to
reduce the number of flat zones in this image, a morphological
leveling is applied on each channel, with markers obtained by an ASF
(Alternate Sequential Filter) of size 1. Then, $\lambda$-flat zones
are computed. Besides, the computation time in our current
implementation with Python is moderate, i.e. a few minutes. However,
note that queues algorithms in C++ are very fast.

\begin{figure}[!ht]
    \centering
    \subfigure[]{\label{\fullpaperid:subfig:1a}\includegraphics[width=0.3\hsize]{\fullpaperpdirectory/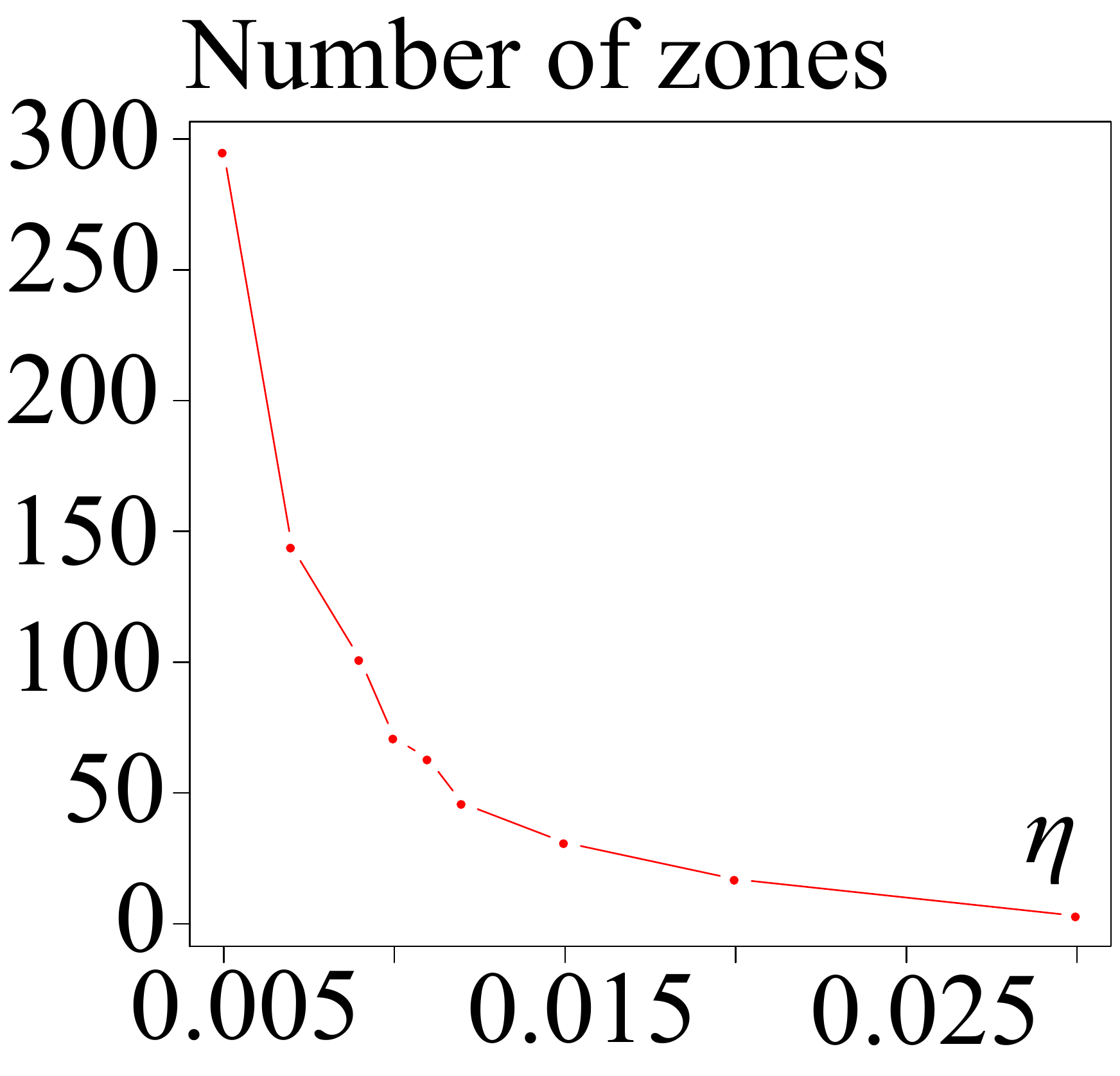}}
    \hspace{0.005\hsize}
    \subfigure[]{\label{\fullpaperid:subfig:1b}\includegraphics[width=0.3\hsize]{\fullpaperpdirectory/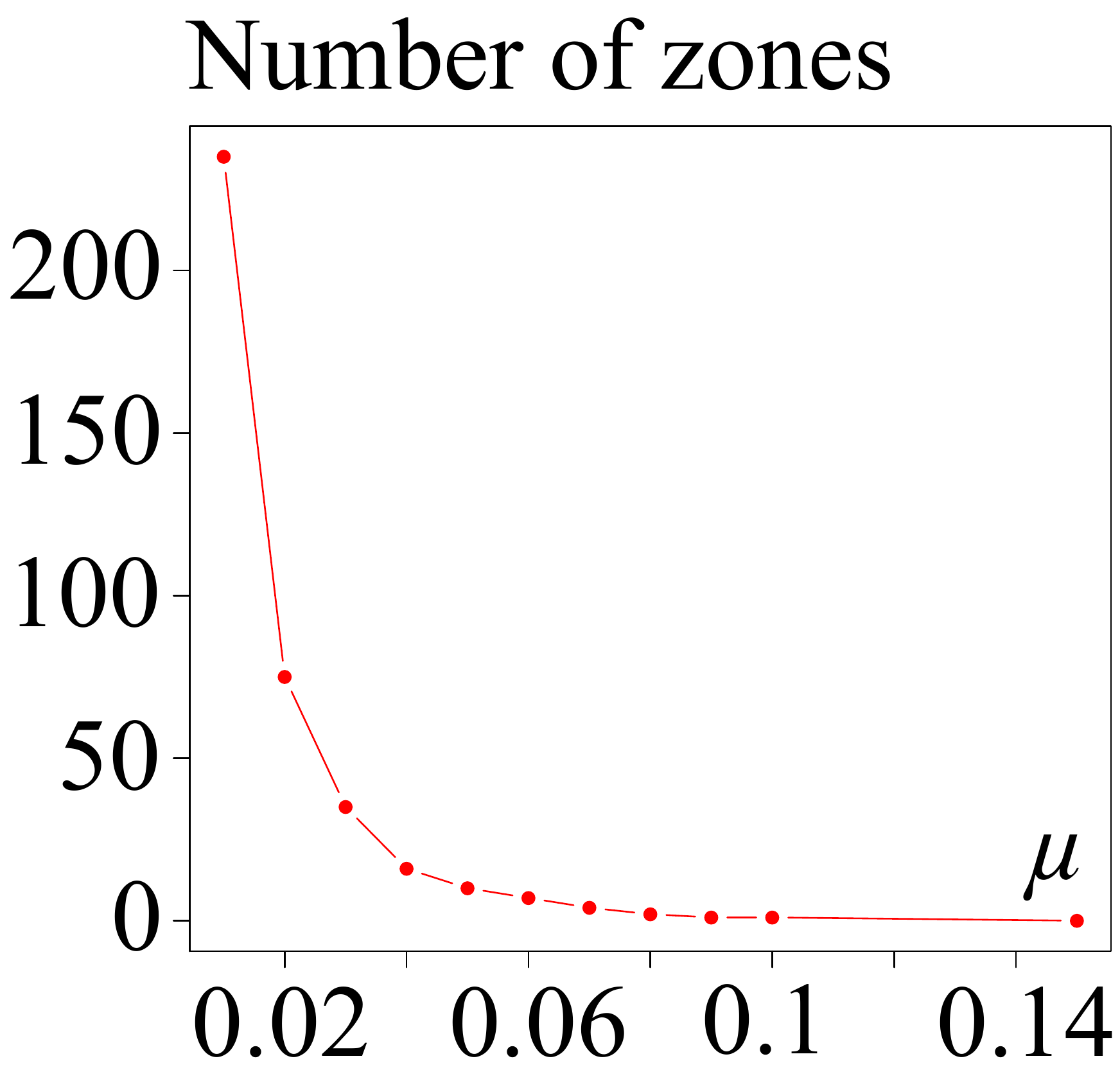}}
    \caption{Variations of the number of $\eta BR$ or $\mu GB$ minus the number of $\lambda FZ$
    versus the parameter $\eta$ or $\mu$ in image "woman face" for $\lambda =
    0.005$ (Chi-squared distance $d_{\chi^2}$).
    (a)~$\eta BR$. (b)~$\mu GB$.}
    \label{\fullpaperid:fig_number_zones_im_woman_face}
\end{figure}

It is important to choose an appropriate distance with respect to
the space of the image. In the spectral initial image of "woman
face" we choose the Chi-squared distance. For $\eta$-bounded regions
and $\mu$-geodesic balls, the number of zones minus the number of
$\lambda$-flat zones is measured. In the
\autoref{\fullpaperid:fig_number_zones_im_woman_face}, we notice
that the number of zones decrease with the parameters $\eta$ or
$\mu$. From figures, we notice that $\eta$-bounded regions are less
sensitive than $\mu$-geodesic balls to small variations on distances
between points. However, the area of $\mu$-geodesic balls is more
controlled than the area of $\eta$-bounded regions. Therefore,
$\eta$-bounded regions are better to find the details and
$\mu$-geodesic balls are better to build smoother zones.

\begin{figure}[!ht]
    \centering
    \subfigure[]{\label{\fullpaperid:subfig:1a}\includegraphics[width=0.18\hsize]{\fullpaperpdirectory/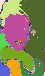}}
    \hspace{0.005\hsize}
    \subfigure[]{\label{\fullpaperid:subfig:1b}\includegraphics[width=0.18\hsize]{\fullpaperpdirectory/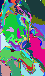}}
    \hspace{0.005\hsize}
    \subfigure[]{\label{\fullpaperid:subfig:1c}\includegraphics[width=0.18\hsize]{\fullpaperpdirectory/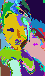}}
    \hspace{0.005\hsize}
    \subfigure[]{\label{\fullpaperid:subfig:1d}\includegraphics[width=0.18\hsize]{\fullpaperpdirectory/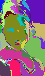}}
    \hspace{0.005\hsize}
    \subfigure[]{\label{\fullpaperid:subfig:1e}\includegraphics[width=0.18\hsize]{\fullpaperpdirectory/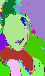}}

    \subfigure[]{\label{\fullpaperid:subfig:1f}\includegraphics[width=0.18\hsize]{\fullpaperpdirectory/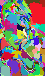}}
    \hspace{0.005\hsize}
    \subfigure[]{\label{\fullpaperid:subfig:1g}\includegraphics[width=0.18\hsize]{\fullpaperpdirectory/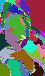}}
    \hspace{0.005\hsize}
    \subfigure[]{\label{\fullpaperid:subfig:1h}\includegraphics[width=0.18\hsize]{\fullpaperpdirectory/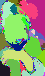}}
    \hspace{0.005\hsize}
    \subfigure[]{\label{\fullpaperid:subfig:1i}\includegraphics[width=0.18\hsize]{\fullpaperpdirectory/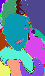}}

    \caption{$\eta$-bounded regions and $\mu$-geodesic balls of image "woman face" for $\lambda =
    0.005$ (Chi-squared distance $d_{\chi^2}$).
    (a)~$\lambda FZ$. $\lambda=0.005$. \newline
    (b)~$\eta = 0.007$. (c)~$\eta = 0.009$. (d)~$\eta=0.011$.
    (e)~$\eta=0.02$.\newline
    (f)~$\mu=0.01$. (g)~$\mu = 0.02$. (h)~$\mu = 0.03$. (i)~$\mu=0.05$.}
    \label{\fullpaperid:fig_eta_BR_mu_GB_im_woman_face}
\end{figure}

\begin{figure}[h]
    \centering
    \subfigure[]{\label{\fullpaperid:subfig:1a}\includegraphics[width=0.18\hsize]{\fullpaperpdirectory/Images/Womanface_leveling_Lambda_0_005_FZ_hyperspectral.png}}
    \hspace{0.005\hsize}
    \subfigure[]{\label{\fullpaperid:subfig:1b}\includegraphics[width=0.18\hsize]{\fullpaperpdirectory/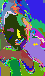}}
    \hspace{0.005\hsize}
    \subfigure[]{\label{\fullpaperid:subfig:1c}\includegraphics[width=0.18\hsize]{\fullpaperpdirectory/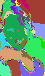}}
    \hspace{0.005\hsize}
    \subfigure[]{\label{\fullpaperid:subfig:1d}\includegraphics[width=0.18\hsize]{\fullpaperpdirectory/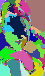}}
    \hspace{0.005\hsize}
    \subfigure[]{\label{\fullpaperid:subfig:1e}\includegraphics[width=0.18\hsize]{\fullpaperpdirectory/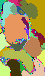}}
    \caption{$\eta$-bounded regions and $\mu$-geodesic balls with an anti-median seed in image
     "woman face" for $\lambda = 0.005$ (Chi-squared distance $d_{\chi^2}$).
    (a)~$\lambda FZ$, $\lambda=0.005$. (b)~$\eta = 0.009$. (c)~$\eta = 0.011$.
    (d)~$\mu=0.02$. (e)~$\mu=0.03$.}
    \label{\fullpaperid:fig_eta_BR_mu_GB_antimedian_im_woman_face}
\end{figure}

Besides, in order to evaluate the influence of choosing the
vectorial median as a reference seed, we have tested the use of the
reverse order for the ascending ordered list based on cumulative
distance. In fact, this order corresponds to the vectorial
anti-median
(\autoref{\fullpaperid:fig_eta_BR_mu_GB_antimedian_im_woman_face}).
Comparing these figure to the zones obtained with the median seed
(\autoref{\fullpaperid:fig_eta_BR_mu_GB_im_woman_face}), we notice
that almost the same zones are obtained. Consequently,
$\eta$-bounded regions and $\mu$-geodesic balls have a small
dependence to the chosen seeds.

Moreover, the same segmentations can be obtained in factor space
using an Euclidian distance because it is equivalent to Chi-squared
distance in image space. We have also computed it, keeping three
factorial axes with relative inertia: 87.2 \%, 10.2 \% and 1.5 \%.
By reducing the volume of data, the computation is more efficient on
3 channels than on 61.

\section{Conclusion and Perspectives}

We have presented two new connected zones: $\eta$-bounded regions
and $\mu$-geodesic balls. They improve the $\lambda$-flat zones,
which deals only with local information, by introducing regional
information. Moreover, these new connections are of second order
because they are built, and included, in the $\lambda$-flat zones
which are already connected. The approach consists in selecting a
sufficiently high parameter $\lambda$ to obtain first a
sub-segmentation. Then, $\eta$-bounded regions or $\mu$-geodesic
balls are built, leading to a segmentation by a top down
aggregation. The $\eta$-bounded regions introduce a parameter
controlling the variations of distance amplitude in the
$\lambda$-flat zones, meanwhile $\mu$-geodesic balls introduce a
parameter to control the size by controlling the cumulative
amplitude inside the $\lambda$-flat zones.

Besides, these two second order connections produce pyramids of
partitions with a decreasing number of regions when the value of
$\eta$ or $\mu$ is increased, until the partition associated to the
last level is equal to the partition defined by the $\lambda$-flat
zones. However, it is important to notice that this pyramid is not
an ordered hierarchy in the meaning that the classes are not ordered
by increasing level.

Furthermore, we have proposed algorithms for the construction of
partitions associated to both new types connections, $\eta$ and
$\mu$, which are based on queues with an ordered selection of seeds
using the cumulative distance.

About the perspectives, we notice that the proposed method does not
solve the problem of small classes of the initial partition of the
$\lambda$-flat zones. However, we can combine our method with the
approaches aggregating smaller regions to these of larger area
\cite{noyel:BrunnerSoille:2005, noyel:CrespoSchafer:1997,
noyel:SalembierGarridoGarcia:1998}. For the future, we are thinking
on more advanced methods in order to select the seeds, and to
determine locally, for each $\lambda$ class, the adapted value of
$\eta$ or $\mu$.

 %%
% BIBLIOGRAPHY
% for details, see ftp://ftp.ams.org/pub/tex/amsrefs/amsrdoc.pdf
%

\end{document}